\documentclass[aoas]{imsart}

\RequirePackage{amsthm,amsmath,amsfonts,amssymb}
\RequirePackage[authoryear]{natbib}

\startlocaldefs

\RequirePackage{amstext}
\RequirePackage{amsthm}
\RequirePackage{amsmath}
\RequirePackage{amsfonts}
\RequirePackage{amssymb}
\RequirePackage{bm}
\RequirePackage{dsfont}
\RequirePackage{mathtools}

\RequirePackage[colorlinks,citecolor=blue,linkcolor=red]{hyperref}

\usepackage{lipsum} 

\usepackage{graphics}
\usepackage{graphicx}
\usepackage{subcaption} 
\usepackage{enumitem}
\usepackage{color}

\usepackage{booktabs}
\usepackage{longtable}
\usepackage{array}
\usepackage{multirow}
\usepackage{wrapfig}
\usepackage{float}
\usepackage{colortbl}
\usepackage{pdflscape}
\usepackage{tabu}
\usepackage{threeparttable}
\usepackage{threeparttablex}
\usepackage[normalem]{ulem}
\usepackage{makecell}
\usepackage{xcolor}

\usepackage{tikz}
\usepackage{extarrows}
\usetikzlibrary{fadings}
\usetikzlibrary{patterns}
\usetikzlibrary{shadows}
\usetikzlibrary{shadows.blur}
\usetikzlibrary{shapes}

\usepackage{lscape}

\usepackage{xparse}

\numberwithin{equation}{section}
\numberwithin{figure}{section}
\numberwithin{table}{section}

\theoremstyle{plain}
\newtheorem{thm}{Theorem}[section]

\newtheorem{prop}{Proposition}[section]

\theoremstyle{definition}

\theoremstyle{definition}

\makeatletter
\def\th@newremark{\th@remark\thm@headfont{\bfseries}} \makeatletter
\theoremstyle{newremark}

\newcommand*{\fullref}[1]{\hyperref[{#1}]{\autoref*{#1} of \nameref*{#1}}}

\makeatletter
\def\namedlabel#1#2{\begingroup
    #2\def\@currentlabel{#2}\phantomsection\label{#1}\endgroup
}
\makeatother

\NewDocumentCommand{\logit}{o}{\IfNoValueTF{#1}{{\rm{logit}}}{{\rm{logit}}\left(#1\right)}}

\NewDocumentCommand{\indicator}{o}{\IfNoValueTF{#1}{\mathds{1}}{\mathds{1}{\lrbrk{#1}}}}

\newcommand*{\pdiff}[2]{\mathchoice{\frac{\partial #1}{\partial #2}}{\partial #1 / \partial #2}{\partial #1 / \partial #2}{\partial #1 / \partial #2}} 
\AtBeginDocument{}

\DeclareMathOperator*{\argmin}{arg\,min}

\NewDocumentCommand{\samplemean}{oo}{\IfNoValueTF{#1}{\IfNoValueTF{#2}{\frac{1}{n}\sum_{i=1}{n}}{\frac{1}{#2}\sum_{i=1}^{#2}}}{\IfNoValueTF{#2}{\frac{1}{n}\sum_{#1}{n}}{\frac{1}{#2}\sum_{#1}^{#2}}}
}
\NewDocumentCommand{\seqsum}{oo}{\IfNoValueTF{#1}{\IfNoValueTF{#2}{\sum_{i=1}{n}}{\sum_{i=1}^{#2}}}{\IfNoValueTF{#2}{\sum_{#1}{n}}{\sum_{#1}^{#2}}}
}

\NewDocumentCommand{\expect}{ e{^} s o >{\SplitArgument{1}{|}}m }{\operatorname{\mathds{E}}\IfValueT{#1}{{\!}^{#1}}\IfBooleanTF{#2}{\expectarg*{\expectvar#4}}{\IfNoValueTF{#3}{\expectarg{\expectvar#4}}{\expectarg[#3]{\expectvar#4}}}}
\NewDocumentCommand{\expectvar}{mm}{#1\IfValueT{#2}{\nonscript\;\delimsize\vert\nonscript\;#2}}
\DeclarePairedDelimiterX{\expectarg}[1]{[}{]}{#1}

\NewDocumentCommand{\prob}{ e{^} s o >{\SplitArgument{1}{|}}m }{\operatorname{\mathds{P}}\IfValueT{#1}{{\!}^{#1}}\IfBooleanTF{#2}{\probarg*{\probvar#4}}{\IfNoValueTF{#3}{\probarg{\probvar#4}}{\probarg[#3]{\probvar#4}}}}
\NewDocumentCommand{\probvar}{mm}{#1\IfValueT{#2}{\nonscript\;\delimsize\vert\nonscript\;#2}}
\DeclarePairedDelimiterX{\probarg}[1]{(}{)}{#1}

\NewDocumentCommand{\expectn}{ s o >{\SplitArgument{1}{|}}m }{\operatorname{\mathds{E}}_{\!n}\IfBooleanTF{#1}{\expectarg*{\expectvar#3}}{\IfNoValueTF{#2}{\expectarg{\expectvar#3}}{\expectarg[#2]{\expectvar#3}}}}
\NewDocumentCommand{\probn}{ e{^} s o >{\SplitArgument{1}{|}}m }{\operatorname{\mathds{P}}_{\!n}\IfValueT{#1}{{\!}^{#1}}\IfBooleanTF{#2}{\probarg*{\probvar#4}}{\IfNoValueTF{#3}{\probarg{\probvar#4}}{\probarg[#3]{\probvar#4}}}}

\makeatletter
\newsavebox{\mybox}\newsavebox{\mysim}
\newcommand{\distas}[1]{\savebox{\mybox}{\hbox{\kern3pt$\scriptstyle#1$\kern3pt}}\savebox{\mysim}{\hbox{$\sim$}}\mathbin{\overset{\text{#1}}{\kern\z@\resizebox{\wd\mybox}{\ht\mysim}{$\sim$}}}}
\makeatother
\newcommand{\defas}{{:=}}

\DeclarePairedDelimiterX{\infdivx}[2]{(}{)}{#1\;\delimsize\|\;#2}

\newcommand*{\cbrk}[1]{\left\{ #1 \right\}}

\newcommand{\lrbrk}[1]{\left(#1\right)}

\newcommand{\norm}[1]{\left\lVert#1\right\rVert}
\NewDocumentCommand{\normF}{o}{\IfNoValueTF{#1}{\norm{\cdot}_F}{\norm{#1}_F}}

\newcommand*{\inv}{^{-1}}

\newcommand*{\tl}[1]{\tilde{#1}}
\newcommand*{\h}[1]{\hat{#1}}

\renewcommand*{\vec}[1]{\bm{#1}}

\NewDocumentCommand{\knl}{o}{\IfNoValueTF{#1}{K}{K\left(#1\right)}}

\DeclareMathAlphabet{\calbnx}{U}{BOONDOX-calo}{m}{n}
\SetMathAlphabet{\calbnx}{bold}{U}{BOONDOX-calo}{b}{n}
\DeclareMathAlphabet{\calbnxb}{U}{BOONDOX-calo}{b}{n}

\usepackage[mathscr]{eucal}

\newcommand{\dsR}{\mathds{R}}

\newcommand{\vphi}{\varphi}

\newcommand{\pos}{^+}

\NewDocumentCommand{\xx}{o}{\IfNoValueTF{#1}{X}{X_{#1}}}
\NewDocumentCommand{\yy}{o}{\IfNoValueTF{#1}{Y}{Y_{#1}}}

\NewDocumentCommand{\f}{o}{\IfNoValueTF{#1}{f}{f_{#1}}
}
\NewDocumentCommand{\fsprt}{o}{\IfNoValueTF{#1}{f_\mu}{f_{\mu, #1}}
}
\NewDocumentCommand{\ff}{so}{\IfBooleanTF{#1}{\IfNoValueTF{#2}{f^*}{f^*_{#2}}}{\IfNoValueTF{#2}{\tl{f}}{\tl{f}_{#2}}}}
\NewDocumentCommand{\hff}{so}{\IfNoValueTF{#2}{\hat{f}}{\hat{f}_{#2}}}

\let\plainsigma\sigma
\RenewDocumentCommand{\sigma}{o}{\IfNoValueTF{#1}{\plainsigma}{\plainsigma_{#1}}}
\NewDocumentCommand{\sff}{so}{\IfBooleanTF{#1}{\IfNoValueTF{#2}{\plainsigma f^*}{\plainsigma f^*_{#2}}}{\IfNoValueTF{#2}{\plainsigma f}{\plainsigma f_{#2}}}}

\newcommand{\lagf}{\mathcal{L}}\newcommand{\dualf}{\mathscr{G}}\newcommand{\rngdualf}{\mathring{\mathscr{G}}}

\NewDocumentCommand{\laga}{so}{\IfBooleanTF{#1}{\IfNoValueTF{#2}{{\vec{a}^*}}{a^*_{#2}}}{\IfNoValueTF{#2}{{\vec{a}}}{a_{#2}}}}
\NewDocumentCommand{\hlaga}{o}{\IfNoValueTF{#1}{{\hat{\vec{a}}}}{\hat{a}_{#1}}}
\NewDocumentCommand{\lagb}{so}{\IfBooleanTF{#1}{\IfNoValueTF{#2}{{\vec{b}^*}}{b^*_{#2}}}{\IfNoValueTF{#2}{{\vec{b}}}{b_{#2}}}}
\NewDocumentCommand{\cc}{so}{\IfBooleanTF{#1}{\IfNoValueTF{#2}{C^*}{C^*_{#2}}}{\IfNoValueTF{#2}{C}{C_{#2}}}}

\NewDocumentCommand{\seqs}{so}{\IfBooleanTF{#1}{\IfNoValueTF{#2}{S^*}{S^*_{#2}}}{\IfNoValueTF{#2}{S}{S_{#2}}}}
\NewDocumentCommand{\hseqs}{so}{\IfBooleanTF{#1}{\IfNoValueTF{#2}{\hat{S}^*}{\hat{S}^*_{#2}}}{\IfNoValueTF{#2}{\hat{S}}{\hat{S}_{#2}}}}
\NewDocumentCommand{\rngseqs}{so}{\IfBooleanTF{#1}{\IfNoValueTF{#2}{\mathring{S}^*}{\mathring{S}^*_{#2}}}{\IfNoValueTF{#2}{\mathring{S}}{\mathring{S}_{#2}}}}

\NewDocumentCommand{\seqsr}{o}{\IfNoValueTF{#1}{S^{(\rho)}}{S^{(\rho)}_{#1}}}
\let\plaintau\tau
\RenewDocumentCommand{\tau}{so}{\IfBooleanTF{#1}{\IfNoValueTF{#2}{\plaintau^*}{\plaintau^*_{#2}}}{\IfNoValueTF{#2}{\plaintau}{\plaintau_{#2}}}}
\NewDocumentCommand{\htau}{so}{\IfBooleanTF{#1}{\IfNoValueTF{#2}{\hat\plaintau^*}{\hat\plaintau^*_{#2}}}{\IfNoValueTF{#2}{\hat\plaintau}{\hat\plaintau_{#2}}}}
\newcommand{\rngtau}{\mathring{\plaintau}}
\let\plainxi\xi
\RenewDocumentCommand{\xi}{so}{\IfBooleanTF{#1}{\IfNoValueTF{#2}{\plainxi^*}{\plainxi^*_{#2}}}{\IfNoValueTF{#2}{\plainxi}{\plainxi_{#2}}}}
\NewDocumentCommand{\hxi}{so}{\IfBooleanTF{#1}{\IfNoValueTF{#2}{\hat\plainxi^*}{\hat\plainxi^*_{#2}}}{\IfNoValueTF{#2}{\hat\plainxi}{\hat\plainxi_{#2}}}}

\let\plainrho\rho
\RenewDocumentCommand{\rho}{so}{\IfBooleanTF{#1}{\IfNoValueTF{#2}{\plainrho^*}{\plainrho^*_{#2}}}{\IfNoValueTF{#2}{\plainrho}{\plainrho_{#2}}}}

\let\plaineta\eta
\RenewDocumentCommand{\eta}{so}{\IfBooleanTF{#1}{\IfNoValueTF{#2}{\plaineta^*}{\plaineta^*_{#2}}}{\IfNoValueTF{#2}{\plaineta}{\plaineta_{#2}}}}
\NewDocumentCommand{\rngeta}{so}{\IfBooleanTF{#1}{\IfNoValueTF{#2}{\mathring{\plaineta}^*}{\mathring{\plaineta}^*_{#2}}}{\IfNoValueTF{#2}{\mathring{\plaineta}}{\mathring{\plaineta}_{#2}}}}

\let\plainmu\mu
\RenewDocumentCommand{\mu}{so}{\IfBooleanTF{#1}{\IfNoValueTF{#2}{\plainmu^*}{\plainmu^*_{#2}}}{\IfNoValueTF{#2}{\plainmu}{\plainmu_{#2}}}}
\NewDocumentCommand{\hmu}{so}{\IfBooleanTF{#1}{\IfNoValueTF{#2}{\hat\plainmu^*}{\hat\plainmu^*_{#2}}}{\IfNoValueTF{#2}{\hat\plainmu}{\hat\plainmu_{#2}}}}
\NewDocumentCommand{\rngmu}{so}{\IfBooleanTF{#1}{\IfNoValueTF{#2}{\mathring{\plainmu}^*}{\mathring{\plainmu}^*_{#2}}}{\IfNoValueTF{#2}{\mathring{\plainmu}}{\mathring{\plainmu}_{#2}}}}
\newcommand{\hparmu}{\hat{\theta}_{\plainmu}}
\NewDocumentCommand{\parmu}{o}{\IfNoValueTF{#1}{\theta_{\plainmu}}{\theta_{\plainmu}^{(#1)}}}

\let\plainnu\nu
\RenewDocumentCommand{\nu}{so}{\IfBooleanTF{#1}{\IfNoValueTF{#2}{\plainnu^*}{\plainnu^*_{#2}}}{\IfNoValueTF{#2}{\plainnu}{\plainnu_{#2}}}}
\NewDocumentCommand{\hnu}{so}{\IfBooleanTF{#1}{\IfNoValueTF{#2}{\hat\plainnu^*}{\hat\plainnu^*_{#2}}}{\IfNoValueTF{#2}{\hat\plainnu}{\hat\plainnu_{#2}}}}
\NewDocumentCommand{\rngnu}{so}{\IfBooleanTF{#1}{\IfNoValueTF{#2}{\mathring{\plainnu}^*}{\mathring{\plainnu}^*_{#2}}}{\IfNoValueTF{#2}{\mathring{\plainnu}}{\mathring{\plainnu}_{#2}}}}
\NewDocumentCommand{\parnu}{o}{\IfNoValueTF{#1}{\theta_{\plainnu}}{\theta_{\plainnu}^{(#1)}}}
\newcommand{\hparnu}{\hat{\theta}_{\plainnu}}

\let\plainzeta\zeta
\RenewDocumentCommand{\zeta}{so}{\IfBooleanTF{#1}{\IfNoValueTF{#2}{\plainzeta^*}{\plainzeta^*_{#2}}}{\IfNoValueTF{#2}{\plainzeta}{\plainzeta_{#2}}}}
\NewDocumentCommand{\hzeta}{so}{\IfBooleanTF{#1}{\IfNoValueTF{#2}{\hat\plainzeta^*}{\hat\plainzeta^*_{#2}}}{\IfNoValueTF{#2}{\hat\plainzeta}{\hat\plainzeta_{#2}}}}
\NewDocumentCommand{\rngzeta}{so}{\IfBooleanTF{#1}{\IfNoValueTF{#2}{\mathring{\plainzeta}^*}{\mathring{\plainzeta}^*_{#2}}}{\IfNoValueTF{#2}{\mathring{\plainzeta}}{\mathring{\plainzeta}_{#2}}}}

 \NewDocumentCommand{\rr}{so}{\IfBooleanTF{#1}{\IfNoValueTF{#2}{\vphi^*}{\vphi^*_{#2}}}{\IfNoValueTF{#2}{\vphi}{\vphi_{#2}}}}
\newcommand{\rngrr}{\mathring{\vphi}}

\NewDocumentCommand{\vcdim}{o}{\IfNoValueTF{#1}{V}{V_{#1}}}

\usepackage{algorithm}
\usepackage{algpseudocode}

\graphicspath{{figures/}}

\usepackage[utf8]{inputenc} \usepackage[T1]{fontenc}    \usepackage{url}            \usepackage{booktabs}       \usepackage{amsfonts}       \usepackage{nicefrac}       \usepackage{microtype}      \usepackage{xcolor}         

\endlocaldefs

\begin{document}

\begin{frontmatter}
\title{Primal--Dual Alternating Neural Learning for Timely Classification with Performance Guarantees}
\runtitle{Timely Decisions with Primal--Dual Neural Learning}

\begin{aug}
\author[A]{\fnms{Jiaming} \snm{Qiu}}, 
\author[A]{\fnms{Yingye} \snm{Zheng}}
\and
\author[A]{\fnms{Ying-Qi}~\snm{Zhao} \thanks{{Corresponding author.}}\ead[label=e3]{yqzhao@fredhutch.org}}
\address[A]{Biostatistics Program, Public Health Sciences Division, Fred Hutchison Cancer Center\printead[presep={ ,\ }]{e3}}

\end{aug}

\begin{abstract}

Timely risk classification is essential in many clinical monitoring settings, where decisions must balance the benefit of classifying patients early for subsequent intervention against the value of observing additional data. Yet most existing statistical and machine-learning methods are designed for fully observed trajectories and offer limited control over key operating characteristics such as sensitivity, specificity, and monitoring cost. We cast the sequential classification problem within a multi-objective optimization framework targeting these three criteria. We characterize the optimal decision rule through a value recursion that quantifies, at each time point, the trade-off between immediate classification and continued monitoring. To estimate the rule from data, we formulate a constrained optimization problem that maximizes specificity while enforcing prespecified sensitivity and monitoring-cost constraints. We then develop an estimation procedure that employs a recurrent neural network to approximate the evolving value processes and a primal--dual updating scheme to satisfy the performance constraints. 
Through simulation studies and an application to continuous glucose monitoring for hypoglycemia risk prediction, we demonstrate that the proposed method yields accurate and timely sequential decision rules that adhere to the desired operating characteristics.
\end{abstract}
 
\begin{keyword}
\kwd{Constrained optimization}
\kwd{Pareto front}
\kwd{Neyman--Pearson classifier}
\kwd{Sequential decision-making}
\end{keyword}

\end{frontmatter}

\section{Introduction}\label{sec:intro}

Many diagnostic and monitoring tasks require sequential decision making, where longitudinal data are used to assess risk and guide timely intervention. In settings such as continuous glucose monitoring (CGM), chronic disease surveillance, and screening programs where sequential measurements accumulate over time, clinicians continually decide whether to act immediately based on the current trajectory or wait for additional observations. Unlike traditional risk prediction models that mainly use baseline information, these practices leverage longitudinally collected data to refine the decision making. Examples include multi-step diagnostic protocols \citep{yu:23}, longitudinal risk prediction using electronic health records \citep{pan:23}, and continuous sensor-based monitoring such as CGM for hypoglycemia risk. These practical problems can be cast as sequential binary classification tasks. However, acting too late increases the monitoring burden and may delay care that could prevent critical events, whereas acting too early may trigger unnecessary alarms or interventions. These trade-offs among sensitivity, specificity, and the cost of continued monitoring are prevalent across clinical and public health applications.

Despite the growing availability of high-frequency sensor data and electronic health records, most existing statistical and machine-learning approaches are not designed for sequential classification under explicit performance constraints. Dynamic treatment regimes and reinforcement-learning methods have gained substantial attention in clinical decision-making in the sequential setup \citep{murp:03, moodie2007demystifying, laber2014dynamic, zhao:15, luckett2019estimating, yu:23, zhou2024estimating, shi2024statistically}, and recent work incorporates safety or risk constraints when learning adaptive treatment policies \citep{laber2014set,liu:24,liu:24-1}. These approaches, however, aim to optimize cumulative reward over sequences of interventions that modify patient trajectories and influence patient outcomes. In contrast, our setting  does not involve altering outcomes through treatment. Instead, the goal is to make timely classification decisions while balancing multiple clinical criteria, such as sensitivity and specificity, and adhering to constraints on monitoring cost. Thus, while methodologically related in spirit, methods on developing dynamic treatment regimes are not directly applicable to constrained sequential classification problems such as ours.

A more closely related line of work is early classification of time series (ECTS), where the goal is to make a class prediction before the entire trajectory is observed. At each time point, these methods decide whether to classify now or wait for more data. Existing ECTS methods typically balance classification accuracy against earliness or measurement cost \citep[see][for a survey]{rena:25}. 
The decision to stop has been guided by prediction reliability \citep{mori:18,scha:20}, anticipated loss reduction \citep{dach:15,ache:21,bils:23}, or likelihood-ratio/posterior-risk stopping boundaries \citep{ebih:21,ebih:24}.
Related ideas have also appeared in longitudinal risk prediction: \citet{pan:23} proposed a reinforced risk prediction method that classifies patients over time while allowing an intermediate ``uncertain'' category for continued monitoring. 
However, these methods often target the accuracy--earliness trade-off. Even when multiple objectives are considered, they typically do not provide direct control of sensitivity and specificity as separate operating characteristics while also treating monitoring cost as a dedicated criterion.\label{mod:ects_literature}

These limitations motivate the need for sequential decision rules that (i) can issue early and interpretable decisions for informing future risks, such that monitoring cost is optimized; (ii) directly satisfy clinically meaningful performance targets, such as high sensitivity to avoid missed events and high specificity to avoid false alarms; and (iii) remain flexible enough to accommodate longer sequences of observations. In this work, we develop a multi-objective optimization (MOO) framework \citep[e.g.,][]{miet:99} for sequential classification. Our approach aligns with a Neyman--Pearson formulation \citep{tong:18}, where we maximize two key design criteria: a sensitivity ensuring protection against missing cases, and a specificity avoiding false positives. We also simultaneously aim to minimize the monitoring cost, which reflects the burden of data collection, delay, or clinical inaction. In applications such as glucose-based hypoglycemia prediction, these objectives translate into fewer missed hypoglycemic events (high sensitivity), fewer false alarms (high specificity), and earlier detection in the sense that low cost corresponds to seeing fewer observations before a decision.

We recast the multi-objective problem as a constrained optimization task that maximizes specificity while enforcing predefined sensitivity and monitoring-cost constraints. From this formulation, we derive a Pareto-optimal sequential decision rule that quantifies, at each time point, the trade-off between acting immediately and gathering additional information. The resulting rule has an intuitive and clinically interpretable form. It recommends providing a decision when the immediate benefit of doing so exceeds the expected benefit of waiting for further data. 

To estimate this rule from data, we propose a learning procedure, termed as Timely Decisions with Primal--dual Alternating Neural Learning (TD-PANL), that uses a recurrent neural network model shared across all decision times to approximate the evolving value processes underlying the decision rule. Neural networks flexibly estimate the evolving conditional risk process and the value processes that dictate when to act. To satisfy the desired sensitivity and monitoring-cost constraints, we introduce a primal--dual alternating scheme that iteratively updates the Lagrange multipliers in the dual and the neural network parameters in the primal. This procedure avoids the need for extensive grid search over multipliers and scales readily to high-dimensional or nonlinear trajectories.  

The remainder of this paper is organized as follows. In \autoref{sec:method}, we formalize the MOO problem and derive the Pareto-optimal sequential rules.
\autoref{sec:esti} provides the details for the TD-PANL method, including the shared model and alternating-step algorithm. \autoref{sec:numexp} reports numerical experiments that investigate the performance of the proposed method and demonstrate the trade-offs among sensitivity, specificity, and cost. The method is also applied to continuous glucose monitoring data to predict hypoglycemia risk. Finally, we provide a discussion in \autoref{sec:diss}. 

\section{Optimal Sequential Decision Rules}\label{sec:method}

Consider a sequence of observations as $\xx = \xx[1:T]$ with $\xx[t] \in \dsR$ for $t = 1, \dots, T$ and a binary outcome $Y \in \cbrk{0, 1}$, with $Y=1$ indicating the presence of an adverse outcome. We observe the data over time and need to determine at each step whether to stop and classify or continue observing. Our goal is to construct optimal sequential rules under competing performance criteria, including specificity, sensitivity, and monitoring cost. In the following, we first define the sequential rules, introduce a constrained Neyman--Pearson formulation that captures the desired performance guarantees, and then derive an explicit form of the optimal rule.

\subsection{Sequential rules and a constrained Neyman--Pearson problem}
A sequential decision rule is a function that maps the history of $\xx[1:t]$ to a decision: ``classify now'' or ``continue'', until the terminal time $T$ when a classification is due. To formalize this, we define $f_t = f_t\lrbrk{\xx[1:t]} \in \dsR$ as the score at time $t$ based on the history $\xx[1:t] = \lrbrk{\xx[1], \dots, \xx[t]}$. For simplicity, we write $f_{1:t}$ as the scores up to time $t$ with the convention that $f_{1:t-1} = 0$ means $f_1, \dots, f_{t-1}$ are all zero. We then define
\begin{equation}\label{def:seq_rule}
    \rr \lrbrk{\xx[1:t]}
    =
    \indicator[\f[1] > 0] +
    \indicator[\f[1] = 0, \f[2] > 0] 
    +
    \cdots +
    \indicator[\f[1:t-1] = 0, \f[t] > 0],
\end{equation}
where $t = 1, \ldots, T$, and $\indicator[\cdot] \in \cbrk{0, 1}$ takes $1$ if the condition is true, and $0$ otherwise.

The decision process can stop before the terminal time $T$ once a non-zero score is received. Denote $\tau = \inf\cbrk{1 \leq t \leq T: f_t \not= 0}$ as the associated \emph{stopping time} at which rule \eqref{def:seq_rule} makes the decision, with $\varphi = 1$ indicating a positive decision at time $\tau$ and $\varphi = 0$ meaning negative decision. Note that the stopping time depends on the observed history $\xx[1:\tau]$, allowing the sequential rule to adapt dynamically to individual trajectories, such as those of different patients.

Let $\cc[t]$ denote the accumulated operational cost of observing up to time $t$. Thus, $0 \leq \cc[1] < \dots < \cc[T]$ and $\cc[\tau]$ is the realized cost of the decision. 
The choice of $\cc$ is application-specific. If observations are equally spaced and similarly burdensome, a natural choice is $\cc[t]=(t-1)/(T-1)$, corresponding to normalized decision time, with zero for deciding at the first time point and one for waiting until the last time point. If late decisions are disproportionately undesirable, one may instead use a nonlinear cost such as $\cc[t]=\lrbrk{(t-1)/(T-1)}^2$. 
If observations differ in burden, e.g., in a diagnostic pathway where later time steps may involve more expensive tests such as imaging or biopsy, one may use $\cc[t]=\sum_{s=1}^t c_s$, with $c_s$ denoting the incremental monetary expense or measurement burden at time $s$. 
Overall, $\cc$ can flexibly encode elapsed time, cumulative measurement burden, or increasing penalty for delayed decision making.
\label{mod:cost}

We consider three characteristics of any sequential decision rules, including sensitivity $\expect{\rr | Y = 1}$, specificity $1 - \expect{\rr | Y = 0}$, and expected cost $\expect{\cc[\tau]}$. These quantities represent competing goals. High sensitivity reduces missed events, but it also leads to higher false positives (lower specificity). Lower cost may correspond to earlier decisions, which might result in less accurate decisions due to the use of less information.

Rather than optimizing a single scalar objective, we adopt a multi-objective optimization (MOO) perspective by identifying the sequential rule that \begin{align}\label{problem:multi_obj}
    \text{maximize } 
        \expect{\rr | Y = 1},\ 
        1 - \expect{\rr | Y = 0},
    \text{ and minimize  }
        \expect{\cc[\tau]}
\end{align}
with expectations taken over $\xx[1:T]$.
This problem seeks to find the optimal trade-off among the aforementioned three competing criteria.

Usually multiple optima of an MOO exist, where none of the objectives can be further improved without sacrificing some other objectives.  Such solutions are called \emph{Pareto optimal}, and their collection is called the \emph{Pareto front} \citep[e.g.,][]{pard:17}. 
Pareto optimality provides a useful interpretation of this MOO formulation. Because cost, sensitivity, and specificity are competing objectives, one should not expect a single sequential rule to dominate across all aspects. A rule on the Pareto front is non-dominated: no other rule can increase sensitivity or specificity, or reduce expected observation cost, without worsening the remaining criteria. Therefore, the Pareto front summarizes the set of efficient operating trade-offs available to the practitioner. Different points on this front then correspond to different viable choices.

To obtain estimable rules that achieve desired operating levels, we adopt the $\varepsilon$-constraints approach. For user-specified targets $0 < \beta < 1$ on sensitivity and $0 < \gamma < 1$ on cost, we find the rule that achieves these targets with the highest specificity:
\begin{align}
    \text{minimize } 
    &\expect{\rr| Y = 0}
    ,
    \nonumber\\
    \text{ subject to }
    &\expect{\cc[\tau]} \leq \gamma
    \text{ and } \expect{\rr | Y = 1} \geq \beta. 
    \label{problem:working}
\end{align} 
This mirrors the Neyman--Pearson paradigm, which maximizes specificity subject to fixed sensitivity, but incorporates cost as an additional operating constraint.

The Lagrangian function of this constraint optimization problem is 
\begin{align}
    \lagf\lrbrk{\f[1:T], a, b} &= 
    \expect{\rr| Y = 0} 
    + a \lrbrk{\expect{\cc[\tau]} - \gamma}
    + b \lrbrk{\beta - \expect{\rr | Y = 1}} \label{def:lagf}
\end{align}
for some non-negative Lagrangian multipliers $(a, b)$.
The Lagrangian multipliers represent the relative importance of the cost, sensitivity, and specificity. For any fixed 
$a$ and $b$, minimizing $\lagf$ corresponds to minimizing a weighted sum of false positives, false negatives, and cost.

\subsection{The optimal decision rule to the Lagrangian}
Define $\mu[t] \defas \expect{Y | \xx[1:t]}$ and $\eta[t] \defas \lrbrk{b p_1\inv + p_0\inv} \mu[t] - p_0\inv$, where $p_1 = \prob{Y=1}$ and $p_0 = 1 - p_1$. Let $\seqs[T] \defas \eta[T]^+ - a \cc[T]$ and recursively
\begin{equation}\label{def:seqS}
    \nu[t] \defas \expect{\seqs[t+1] | \xx[1:t]}
    , \quad
    \seqs[t] \defas \max\lrbrk{\eta[t]^+ - a\cc[t], \nu[t]}
\end{equation}
for $t = 1, \dots, T-1$. Here $\eta[t]^+ \defas \max(\eta[t], 0)$. 
See \citet[Supplement, Section S1]{tdpanl:supp} for the proof of the following proposition.

\begin{prop}\label{prop:dual}
For any fixed Lagrange multipliers $a, b \geq 0$, the rule minimizing the Lagrangian \eqref{def:lagf} is
\begin{equation}\label{eq:opt_f_form}
        \ff[t](\xx[1:t]; a, b) = 
        \begin{cases}
            1, &\text{ if } \eta[t] > 0 \text{ and } \zeta[t] > \nu[t]
            , \\
            0, &\text{ if } \nu[t] \geq \zeta[t]
            , \\
            -1, &\text{ if } \eta[t] \leq 0 \text{ and } \zeta[t] > \nu[t],
        \end{cases}
    \end{equation}
    with $\zeta[t] \defas \eta[t]\pos - a\cc[t]$ and $\nu[t]$ as defined above. 
\end{prop}
    Here, $\ff[t] = 1$ denotes stopping with a positive decision, $\ff[t] = 0$ denotes continuation, and $\ff[t] = -1$ denotes stopping with a negative decision. This is consistent with \eqref{def:seq_rule}, where the indicator sum in \eqref{def:seq_rule} records whether the stopping decision is positive, while the stopping time is determined by the first $t$ such that $\ff[t]\neq 0$. Thus, a negative value of $\ff[t]$ corresponds to a negative termination  $\rr=0$.
Particularly, $\eta[t]$ is the evidence in favor of a positive classification, and its sign dictates whether stopping at time $t$ leads to a positive or negative decision.
    The quantity $\zeta[t] = \eta[t]^+ - a\cc[t]$ is the net immediate benefit of stopping at time $t$, after accounting for the monitoring cost already incurred. The quantity $\nu[t]$ is the expected benefit of continuing to observe the trajectory and making the decision later.
    Thus, the rule stops when the immediate benefit $\zeta[t]$ exceeds the continuation value $\nu[t]$, and otherwise continues monitoring.
    Later, in \autoref{sec:cgm}, we plot $\eta[t]$, $\nu[t]$, and $\zeta[t]$ along observation trajectories to illustrate how these quantities govern the decision process in practice.
\label{mod:explain_form}

\subsection{Dual characterization and Pareto optimality}\label{sec:pareto_optim}
The optimal rule in \autoref{prop:dual} solves the Lagrangian \eqref{def:lagf} for any fixed pair of multipliers $(a,b)$.
A naive way to approximate the full Pareto front would be to compute the rule \eqref{eq:opt_f_form} over a grid of $(a,b)$ values and record the resulting specificity, sensitivity, and cost.
    However, the multipliers do not directly control these interpretable operating characteristics. A regular grid over $(a,b)$ therefore need not translate into a regular coverage of the Pareto front: it may produce many similar rules in some regions while leaving other regions sparsely covered (see Figure~S2.4 of the Supplement).
\label{mod:motivate_dual}

A more principled approach is to examine the dual function of the Lagrangian function $\lagf$, defined as $\dualf(a, b) \defas \inf_{\f[1:T]} \lagf\lrbrk{\f[1:T]; a, b}$, which, by \autoref{prop:dual}, is 
\begin{equation}\label{eq:dualf}
    \dualf(a, b) = b \beta - a \gamma - \expect{\seqs[1]}.
\end{equation}
For any fixed pair $(a,b)$, minimizing the Lagrangian corresponds to finding the best rule under a particular weighted trade-off among false positive rate, sensitivity, and monitoring cost. The dual function records the optimal value achievable under this trade-off. By construction, the dual function $\dualf$ is a lower bound on the minimal value of the constrained problem. Maximizing $\dualf(a,b)$ over nonnegative $(a,b)$ therefore finds the trade-off weights that most tightly support the constrained optimum.
We establish below that, the maximum of the dual function equals the minimum of the primal objective, i.e., the primal and dual problems have no duality gap. Consequently, the point at which $\dualf(a,b)$ achieves its maximum corresponds exactly to the Lagrange multipliers that enforce the sensitivity and cost constraints.

Formally, we consider the dual problem as
\begin{equation}\label{problem:dual}
    \text{maximize $\dualf(a, b)$ for $a, b\geq 0$}.
\end{equation}
We show that, there always exists some dual optimal $a^*, b^*$ with no duality gap, that is, the $\lagf$ and $\dualf$ equal. At the optimal multipliers $a^*, b^*$, the optimal rule $\rr*$  satisfies the sensitivity and cost constraints in \eqref{problem:working} with equality and achieves the lowest possible false positive rate among all rules satisfying those constraints. This also explains the connection to Pareto optimality. If another rule could reduce false positive rate without decreasing sensitivity or increasing cost, it would contradict the optimality of $\rr^*$ for the constrained problem. Therefore, $\rr*$ is Pareto optimal for the MOO. 
These results are summarized in the following theorem.

\begin{thm}\label{thm:pareto}
    Suppose for $t = 1, \dots, T$, the $\xx[t]$ has a continuous density, and $\xx[1:t] \mapsto \mu[1:t]$ are continuous functions bounded away from zero, then for any $\beta, \gamma \in (0, 1)$ there exists a $\rr*$ whose scores take the form of \eqref{eq:opt_f_form} such that
    \begin{enumerate}
        \item $\rr*$ solves the constrained problem \eqref{problem:working} with equality;
        \item $\rr*$ is Pareto optimal for the MOO problem \eqref{problem:multi_obj}.
    \end{enumerate}
\end{thm}
One key assumption here is that $\mu[t]$ is a continuous function of $\xx[1:t]$ for all $t$. The continuity assumption ensures that sensitivity varies continuously with the decision threshold, making it possible to attain any desired sensitivity level without randomization. Essentially, the optimal rule compares the value of stopping versus continuing, providing a generalization of the Neyman--Pearson paradigm to sequential settings. We refer to Section S1 of the Supplement for proof.

In this sense, the theorem links the constrained problem and the original MOO formulation.
Each choice of constraint levels $(\beta,\gamma)$ corresponds to a particular point on the Pareto front. Solving the dual problem for this pair yields the Lagrange multipliers $(a^*, b^*)$ that enforce the constraints and the corresponding optimal rule. As $(\beta,\gamma)$ vary over their feasible ranges, the resulting solutions lead to the full collection of triples that cannot dominate each other. Thus, the $\varepsilon$-constraint formulation, combined with the dual characterization, provides a complete parametrization of the Pareto front. 
This optimal rule structure guides the estimation strategy developed in \autoref{sec:esti}.

\section{Estimating the Optimal Rule via Primal-Dual Alternating Neural Learning}\label{sec:esti}

\autoref{sec:method} provides the theoretical form of the optimal sequential rule. For fixed Lagrange multipliers $(a,b)$, the optimal rule depends on two processes, $\mu[t] = \expect{Y | \xx[1:t]}$ and $\nu[t] = \expect{\seqs[t+1] | \xx[1:t]}$. If $\mu[t]$ and $\nu[t]$ were known, the optimal sequential rule as defined in \eqref{eq:opt_f_form} can be obtained. In real-world applications, these quantities must be estimated from data. We introduce a Timely Decisions with Primal--dual Alternating Neural Learning (TD-PANL) approach that uses neural networks as flexible estimators for $\mu[t]$ and $\nu[t]$, and alternates between  
\begin{enumerate}
    \item primal updates on model parameters, which refine the estimate of $\nu[t]$;
    \item dual updates, which adjust $(a, b)$ so that the sensitivity and cost constraints are satisfied. 
\end{enumerate}

This approach combines the constrained Neyman--Pearson formulation of \autoref{sec:method} with neural network approximations of the value processes $\mu[t]$ and $\nu[t]$. 
The process $\mu[t]$ is trained once to estimate the conditional class probability process, while the process $\nu[t]$ is updated iteratively to approximate the value driving the stopping rule. The dual parameters $(a,b)$ are updated in parallel to enforce sensitivity and cost constraints.

In particular, we will employ Recurrent Neural Networks (RNNs) in estimating $\mu[t]$ and $\nu[t]$. Designed for sequence-to-sequence learning, RNNs compute the output at time $t+1$ as a function of the new observation $\xx[t+1]$, and the hidden state (or output) carried over from time $t$. This recursive evaluation eliminates the need to stack a list of time-specific models. Moreover, the resulting output at time $t$ will depend only on the observed history $\xx[1:t]$. 
Combined with the well-known flexibility and expressiveness of neural networks, RNNs form an ideal estimator class for approximating the processes $\mu[1:T]$ and $\nu[1:T-1]$ required for our method.

\subsection{Estimation of $\mu[t]$ and $\nu[t]$} 
We first estimate the quantities in \autoref{prop:dual}, assuming the Lagrangian multipliers $a, b$ are fixed. 

For each $t$, the risk process $\mu[t] = \expect{Y | \xx[1:t]}$ is the minimizer of the squared error $$\sum_{t=1}^T \expect{\lrbrk{g_t - \yy}^2}$$ over all history-dependent functions $g_{1:T}$. We therefore can estimate $\mu[t]$ by minimizing the empirical analogue of this loss. Estimation of $\nu[t]$ is more challenging because it depends on its future through $\seqs[t+1]$ as defined in \eqref{def:seqS}. Using the recursion form of $\nu[t]  = \expect{\seqs[t+1] | \xx[1:t]}$, we employ a temporal difference learning approach, where the process $\nu[1:T-1]$ is obtained by minimizing
\begin{align}
    \sum_{t=1}^{T-1} \expect{\lrbrk{g_t - \seqs[t+1]}^2}
    \label{eq:nu_loss_population}
\end{align}
which fits all $\nu[t]$ simultaneously. This temporal difference-based loss function aggregates across all time $t = 1, \dots, T$ without backward recursion or fitting a specific model at each time point.

We estimate both $\mu[t]$ and $\nu[t]$ using RNNs.
Denote $\rngmu\lrbrk{\xx[1:t], \parmu}$ and $\rngnu\lrbrk{\xx[1:t], \parnu}$ as the RNN models for $\mu[t]$ and $\nu[t]$ respectively. Substituting these estimates, together with the sample class proportions $\h p_0$ and $\h p_1$, into the definitions in \autoref{sec:method} yields the plug-in quantities $\rngeta, \rngzeta, \rngseqs$, $\rngtau$, and $\rngrr$.
The least squared errors for $\mu[t]$ and $\nu[t]$ are approximated by the corresponding empirical loss functions
\begin{align}
    Q_{\mu}\lrbrk{\parmu} &= \sum_{t = 1}^T \expectn{\lrbrk{\rngmu\lrbrk{\xx[1:t], \parmu} - Y}^2}
    , \nonumber\\
    Q_{\nu}\lrbrk{\parnu | \parmu, a, b} &= \sum_{t = 1}^{T-1} \expectn{\lrbrk{\rngnu\lrbrk{\xx[1:t], \parnu}- \rngseqs[t+1]}^2},
    \label{eq:mnu_loss}
\end{align}
where $\expectn{\cdot}$ denotes the empirical expectation.
The objective $Q_{\mu}$ corresponds to a many-to-one classification problem that uses the entire sequence $\xx[1:T]$ to predict $Y$.  
We first obtain $\hparmu \in \argmin Q_{\mu}$. For fixed $(a,b)$, we compute $\hparnu \in \argmin Q_{\nu}\lrbrk{\cdot|\hparmu, a, b}$. Plugging-in the fitted networks, denoted as $\hmu = \rngmu\lrbrk{\xx[1:t], \hparmu}, \hnu = \rngnu\lrbrk{\xx[1:t], \hparnu}$, into \eqref{eq:opt_f_form} provides the estimated optimal rule for the given Lagrangian multiplier $a, b$.

\subsection{Alternating updates for model parameters and Lagrange multipliers}\label{sec:esti_laga}
A key intermediate result in the proof of \autoref{thm:pareto} is the following characterization of the dual gradient.
\begin{prop}\label{prop:grad_dual}
    The gradient of the dual function is
    $\pdiff{\dualf}{a} = \expect{\cc[\tau]} - \gamma$
    and
    $\pdiff{\dualf}{b} = \beta - \expect{\tl{\rr} | Y = 1}$,
    where $\tl{\rr}$ is the rule \eqref{def:seq_rule} with scores $\ff = \ff(\xx; a, b)$ of \eqref{eq:opt_f_form} plugged-in. 
\end{prop}
This shows that the dual gradient consists of performance gaps, that is, the difference between the achieved and desired cost and sensitivity. The empirical gradients can therefore be approximated by substituting empirical performance measures of the plug-in rule:
\begin{equation}\label{eq:dualf_grad}
    \pdiff{\rngdualf}{a}\lrbrk{a, b | \hparmu, \hparnu} \approx \expectn{\h\cc} - \gamma
    \text{, and }
    \pdiff{\rngdualf}{b}\lrbrk{a, b | \hparmu, \hparnu} \approx \beta - \expectn{\h{\rr} | Y = 1}.
\end{equation}
These expressions have an intuitive interpretation, where $a$ should be increased if the classifier has too much cost, and $b$ should be increased if the sensitivity is lower than $\beta$. Hence, any gradient-based optimizer can be applied to maximize $\rngdualf\lrbrk{a, b | \hparmu, \hparnu}$, provided we update $\nu$ as the dual variables change.

A direct gradient-based approach would require refitting $\nu$ to minimize $Q_{\nu}$ for every update of $(a,b)$, which is computationally intensive.
However, modern gradient ascent uses small step sizes, controlled by learning rates and gradient clipping. Therefore, the change in $(a,b)$ from one iteration to the next is typically modest. Consequently, the corresponding drift in the optimal $\nu$-network is also small, and it suffices to patch $\hparnu$ with a single gradient step on $Q_{\nu}$ rather than re-estimating it from scratch. This leads to an alternating dual-ascent / primal-descent scheme: 
\begin{enumerate}
    \item Dual ascent: update $(a,b)$ using the empirical gaps in cost and sensitivity.
    \item Primal descent: update $\theta_{\nu}$ with one step of gradient descent on $Q_{\nu}$. 
\end{enumerate}
This alternating structure forms the basis of \autoref{alg:nu_laga}.

\begin{algorithm}
    \caption{Optimizing $\parnu$, $a$, and $b$ given $\beta, \gamma$ for $K$ iterations with $\delta$ tolerance and learning rate $r$.}\label{alg:nu_laga}
    \begin{algorithmic}
        \State Obtain $\hparmu \gets \argmin Q_{\mu}$, then initialize $a^{(0)}$ and $b^{(0)}$, set $k = 0$.
        \State $\parnu[0] \gets \argmin Q_{\nu}\lrbrk{\cdot| \hparmu, a^{(0)}, b^{(0)}}$.
        \While{$k \leq K$ and $\norm{\partial \rngdualf} > \delta$}
        \State Lagrangian update: $\lrbrk{a^{(k+1)}, b^{(k+1)}} \gets \lrbrk{a^{(k)}, b^{(k)}} + r \partial {\rngdualf(a, b | \hparmu, \parnu[k])}$;
        \State $\nu$-update: $\parnu[k+1] \gets \parnu[k] - r \partial {Q_{\nu}(\parnu | \hparmu, a^{(k+1)}, b^{(k+1)})}$;
        \State $k \gets k + 1$.
        \EndWhile
    \end{algorithmic}
\end{algorithm}

    The Lagrangian multipliers $(a,b)$ are updated using the empirical gaps in cost and sensitivity, so that the induced rule moves toward the target levels $(\beta, \gamma)$. The update of $\parnu$ is nested within this process, allowing the rule and the multipliers to be learned jointly from the training data.
    All parameter updates are based on the training data, while a separate validation set can be used to monitor performance and stop training.
    In particular, training is stopped when either the maximum number of iterations $K$ is reached or the norm of the dual gradient $\norm{\partial \rngdualf}$ falls below a prespecified tolerance $\delta$. By \autoref{prop:grad_dual}, the latter means that the empirical cost and sensitivity gaps are both sufficiently small. In our implementation, $K$ was chosen sufficiently large so that training typically stopped by the performance rather than by hitting the iteration limit.
\label{mod:training}

In our implementation, the training samples are representative, and $p_0$ and $p_1$ are estimated by the sample class proportions.
However, in applications where one class is rare, class-dependent sampling, such as upsampling the minority class or downsampling the majority class, may be used to improve class representation. Such sampling changes the class proportions in the training sample.  Supplement Section S3 describes an extension to class-dependent sampling, including corrections for estimating $\mu[t]$, $\nu[t]$, and the dual updates.\label{mod:class_balancing}

\section{Numerical Experiments}\label{sec:numexp}

This section evaluates the empirical performance of the proposed TD-PANL method using two complementary examples. We begin with a synthetic study, which mimics the sequential monitoring of longitudinal trajectories in practice. These examples have increasing complexity to illustrate how sequential structure, nonlinearity, and signal-to-noise ratio affect the estimation of the value processes. 
We then apply TD-PANL to a real continuous glucose monitoring dataset \citep[CGM,][]{wein:15} to evaluate early hypoglycemia detection, where timely decisions are clinically important. Across settings, we assess whether the learned rules satisfy the desired sensitivity and cost constraints. 

For simplicity, across all experiments we set the cost function to $\cc[t] = (t - 1)/(T - 1)$, yielding a unified cost scale of $[0,1]$ regardless of the sequence length $T$. 
For plug-in estimators for $\mu$ and $\nu$, the proposed method used two separate gated recurrent units \citep[GRUs;][]{chun:14}, each followed by a fully connected linear post-processing layer. The resulting decision rule contained approximately 2,000 to 3,000 trainable parameters depending on $T$. Details of the network architecture are provided in Section S2 of the Supplement.

\subsection{
    Synthetic Data Examples}\label{sec:simu_ar}

\subsubsection{Data Generation}
We begin with simulated studies designed to mimic sequential monitoring of longitudinal measurements. The data are generated under three temporal mechanisms: a Markov process, a probit model, and a bi-modal mixture model. These settings vary in temporal and signal complexity, thereby probing different aspects of learning the processes $\mu[t]$ and $\nu[t]$. 
Particularly, we generate $\xx[1:T]$ as a stationary autoregressive time series with standard Gaussian innovations. We draw $\yy$ following $\prob{\yy=1 | \xx[1:T]} = h\lrbrk{V}$, where $V$ is a linear combination of $\xx[1:T]$ and $h:\dsR \to [0, 1]$ maps this score to probability. We varied the auto-regressive parameters, the link $h$, and the coefficients of $V$ to create the three simulation settings, where the marginal event rates $\prob{\yy = 1}$ are both 0.5 for the Markov and probit settings, and 0.27 for the bi-modal setting. The detailed specification can be found in Section S2.1 of the Supplement.

The sequence length is set to $T=5$, allowing us to compute the true optimal by numerical integration, which becomes prohibitive for larger $T$. The proposed method itself does not require this restriction. 
Each setting was repeated 50 times. In each replicate, we generated $10^4$ and $10^5$ series for training and testing respectively. 
An additional 2,500 validation series were generated for performance monitoring and stopping training.
A grid of sensitivity and cost constraints $\beta, \gamma \in 0.1, \dots, 0.9$ were considered. 
\subsubsection{Methods for Comparison}
For the simulated settings, we numerically computed the optimal rule using the known data-generating mechanism, providing a theoretical benchmark for the best achievable specificity--cost trade-off under the imposed sensitivity constraint.

We compared the proposed method to a simple myopic rule, which bases its decision solely on the currently observed data at a pre-specified time $t_0$ and ignores any potential future information. Operationally, this places a Neyman--Pearson classifier \citep{tong:18} at $t_0$, so the myopic rule can be constructed directly from the estimated probabilities $\hmu[t_0]$. Because the myopic rule does not learn a data-dependent stopping time, its cost is fixed at $\cc[t_0]$.

    We additionally compared the proposed method with FIRMBOUND \citep{ebih:24}, an ECTS method closely related to our optimal-stopping formulation. Additional details on FIRMBOUND are provided in Section S2.2 of the Supplement. 
    Like TD-PANL, FIRMBOUND treats early classification as a finite-horizon stopping problem and compares immediate stopping risk with expected continuation risk. However, FIRMBOUND performs this recursion using the posterior risk $\mu[t]$ as a low-dimensional summary of the observed history, whereas TD-PANL conditions on the full trajectory $\xx[1:t]$. Additionally, FIRMBOUND optimizes a prespecified weighted combination of classification error and observation cost, rather than directly targeting sensitivity and cost constraints. 
    Consequently, identifying operating points with desired sensitivity and cost levels requires tuning over a grid of weight parameters and selecting rules whose realized operating characteristics match the target constraints. Particularly, we first performed a broad coarse grid search over its weight parameters on $[0,5]\times[0,5]$ to identify the range producing sensitivity near 0.9. We then refined the search using a 20-by-20 grid over the smaller region $[0.01, 1]\times[1,2.3]$. Finally, we used post hoc matching: for each cost constraint considered, we selected FIRMBOUND rules whose realized sensitivity and cost were comparable, and then compared specificity.

\subsubsection{Results}

\begin{figure}[!h]
    \centering
    \begin{subfigure}{\textwidth}
        \centering
        \includegraphics[width=0.9\linewidth]{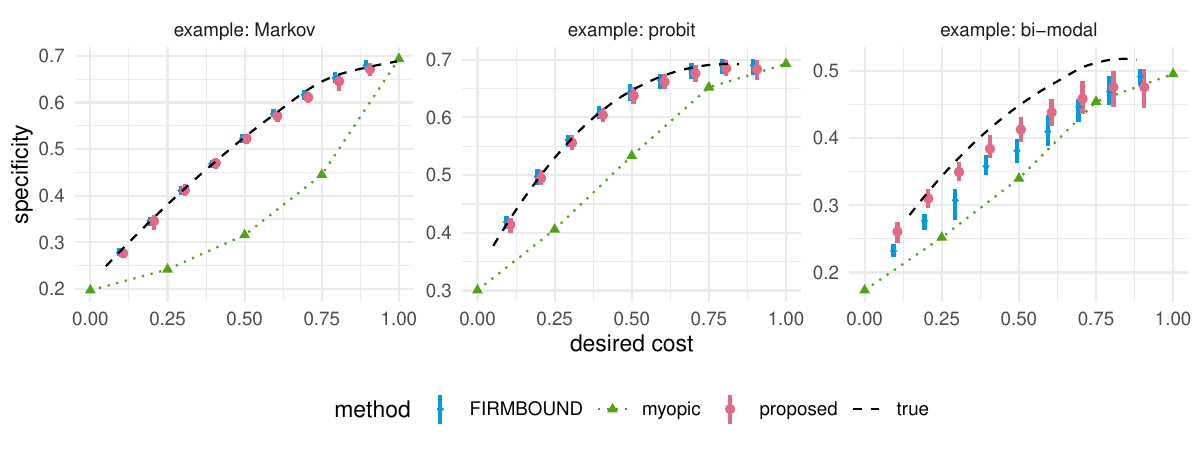}
        \caption{
            Specificity achieved by different methods at varying monitoring-cost levels.
        }\label{fig:simu_pertpr_fpr}
    \end{subfigure}
    \\
    \begin{subfigure}{\textwidth}
        \centering
        \includegraphics[width=0.9\linewidth]{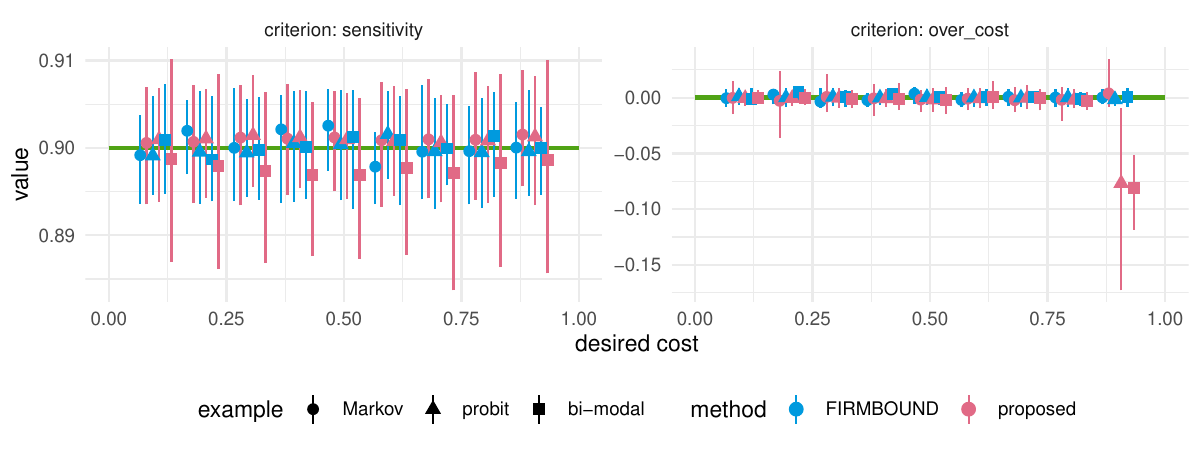}
        \caption{
            Constraints on sensitivity (left) and cost (right) are satisfied by the proposed method.
        }\label{fig:simu_pertpr_cons}
    \end{subfigure}
    \caption{
        Performance of the proposed sequential decision rules, FIRMBOUND competitor, the myopic baselines, and the true optimal, under a constraint of sensitivity at $\beta = 0.9$ and cost over $\gamma = 0.1, \dots, 0.9$. 
        The top panels illustrate the specificity--cost trade-off under a sensitivity constraint. The bottom panels inspect whether the constraints were satisfied, where the over\_cost (the bottom right panel) is defined by the achieved cost minus the desired. 
        The solid horizontal lines also represent the myopic baselines, which satisfy the sensitivity and cost constraints by construction.
        The means (dots) and 90\% intervals (bars) over 50 repeated experiments are reported. 
    }\label{fig:simu_pertpr}
\end{figure}

To illustrate the specificity--cost trade-off, \autoref{fig:simu_pertpr} constrains the sensitivity constraint at $\beta=0.9$ and varies the cost constraint over $\gamma=0.1,\ldots,0.9$. As demonstrated in \autoref{fig:simu_pertpr_fpr}, the proposed method nearly coincides with the true optimal Pareto frontier in the Markov and probit settings, because both $\mu[t]$ and $\nu[t]$ are relatively easy to learn. In the bi-modal setting, TD-PANL still  recovers the overall trade-off  but exhibits a relatively larger gap from the optimum due to the increased difficulty of estimating $\mu[t]$ under a non-monotone, bimodal link $h$. Supplementary Section S2.1 shows that when the true $\mu[t]$ is supplied, the proposed method recovers the optimal frontier. \autoref{fig:simu_pertpr_cons} further shows that the sensitivity and cost constraints are satisfied throughout the considered operating range. These results indicate that TD-PANL can attain performance close to the optimal stopping rule while remaining computationally feasible.

The resulting frontier also provides insight into the value of additional observations. Notably, the gain in specificity diminishes as cost increases, especially in the probit and bi-modal examples, indicating that once specificity is near its maximal value, additional observations yield limited gains. This is also reflected in the excess cost panel (bottom right of \autoref{fig:simu_pertpr_cons}) where the rightmost bars are negative. Together, these patterns suggest an oracle-like behavior, in the sense that the learned rule often reaches a decision using only a shorter sequence of observations while achieving performance close to that attainable using the full trajectory. 

The myopic rules' performance was evaluated directly on the testing set using an ROC analysis of the predicted probabilities, so they always satisfy the requested sensitivity constraints. Relative to the myopic benchmark, TD-PANL consistently achieves higher specificity across the three simulation settings. These results demonstrate the benefit of adaptive stopping, compared to a fixed-time classification.

    Compared with FIRMBOUND, the proposed TD-PANL achieved comparable specificity in the Markov and probit settings  while maintaining similar sensitivity and observation cost. In the bi-modal setting, TD-PANL achieved higher specificity, with the advantage most pronounced under stringent cost constraints. The larger gap in the bi-modal setting may reflect that $\mu[t]$ does not summarize the stopping problem as effectively as in the other two settings. In the Markov and probit examples, the Bellman recursion \eqref{def:seqS} can be written using $\mu[t]$ alone rather than the full history $\xx[1:t]$. In contrast, in the bi-modal setting, the same value of $\mu[t]$ may arise from different histories that imply different future evolution. As a result, conditioning only on $\mu[t]$ may discard information relevant to the stopping decision.
\label{mod:ects}

\begin{figure}[!h]
    \centering
    \includegraphics[width=0.9\linewidth]{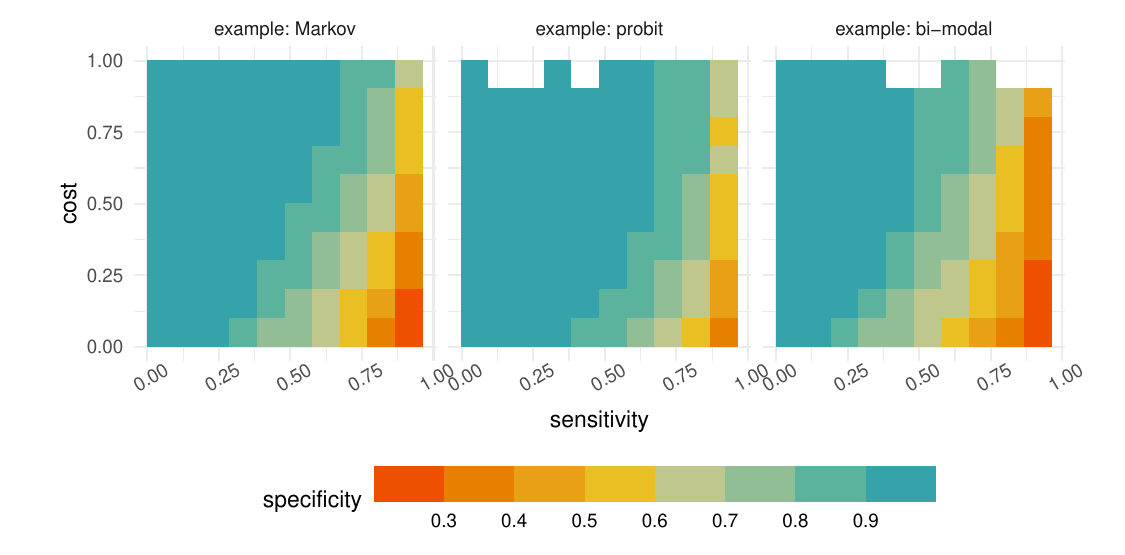}
    \caption{
        The operating characteristic heatmaps of the sequential rules illustrates the three-way trade-off among specificity, sensitivity, and cost. Each tile represents the binned and averaged performance over 50 repeated experiments.
    }\label{fig:simu_pareto}
\end{figure}

    Additionally, we examined the full three-way trade-off among specificity, sensitivity, and cost in \autoref{fig:simu_pareto}. We solved the working problem over a grid of sensitivity and cost constraints spanned by $\beta, \gamma \in 0.1, \dots, 0.9$. Each fitted rule was evaluated on the test set, producing realized sensitivity, cost, and specificity. 
In the heatmaps, the fitted rules were then grouped into bins according to their realized sensitivity and cost, and the color of each tile gives the average realized specificity within that bin. The figure therefore provides a visual summary of the three-way trade-off.
Across the three examples, high specificity is maintained over a broad region of the sensitivity--cost plane. The loss of specificity is concentrated in the lower-right region requiring both high sensitivity and low cost, where the rule must stop early while still identifying most positive cases. Conversely, when the sensitivity requirement is relaxed or more observations are allowed, high specificity is easier to maintain. These results indicate that favorable operating characteristics can be achieved across a wide range of sensitivity and cost targets, rather than only at the fixed sensitivity level considered in \autoref{fig:simu_pertpr}. 

    Figure~S2.3 in the Supplement re-expresses the operating characteristics shown in \autoref{fig:simu_pareto} in terms of the corresponding learned multipliers. The highly nonlinear relationship between the multipliers and the resulting operating characteristics provides additional motivation for the adaptive dual updates used by the proposed method.

Finally, we conducted a sensitivity analysis to assess the robustness of the simulation results to the choice of neural network architecture, model depth, and hidden dimension. We considered a deeper GRU model, a wider GRU model, and alternative models including a vanilla RNN, an LSTM, and a Transformer encoder with a causal mask. Overall, the empirical conclusions were broadly stable across these choices. See Section S2.3 of the Supplement for details.

\clearpage

\subsection{Early Detection of Hypoglycemia through Continuous Glucose Monitoring}\label{sec:cgm}

We applied the proposed sequential decision-rule framework to continuous glucose monitoring (CGM) data from a cohort of older adults ($\geq$ 60 years) with long-standing Type 1 diabetes \citep{wein:15}\footnote{Data accessed from \url{https://public.jaeb.org/datasets/diabetes}, \textit{Severe Hypoglycemia in Older Adults with Type 1 Diabetes: A Study to Identify Factors Associated with the Occurrence of Severe Hypoglycemia in Older Adults with T1D}.}. The original cohort included 201 participants, with 101 cases who experienced at least one severe hypoglycemic episode in the prior 12 months and 100 controls without such events in the preceding three years. All participants wore blinded CGM for up to 14 days, during which glucose was recorded without additional intervention, capturing natural within-person variation.

Severe hypoglycemia poses significant risks, especially for older adults, including falls, cardiovascular events, and cognitive impairment.
Early detection of impending hypoglycemia enables timely interventions, such as carbohydrate intake or insulin adjustment, which is easier to implement and more effective than treating an event once it has already occurred.
Our task is to predict impending hypoglycemia using preceding CGM.

\begin{figure}
    \centering
    \begin{subfigure}{\textwidth}
    \includegraphics[width=0.9\linewidth]{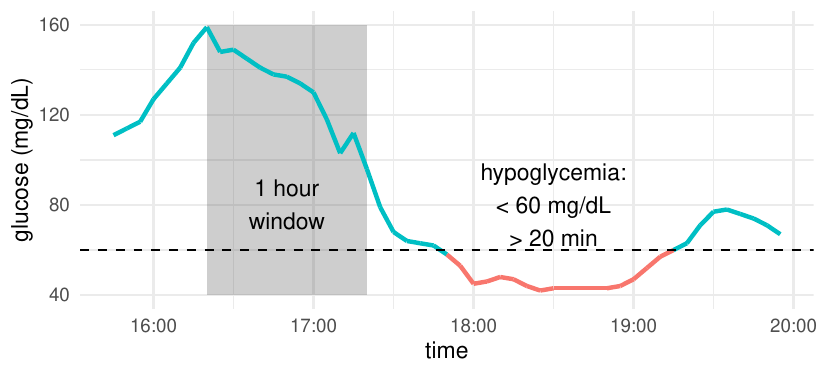}
    \caption{
        An example one-hour monitoring window (shaded), followed by a  hypoglycemia episode within 30 minutes.
    }\label{fig:cgm_window}
    \end{subfigure}\\
    \begin{subfigure}{0.475\textwidth}
\includegraphics[width=\linewidth]{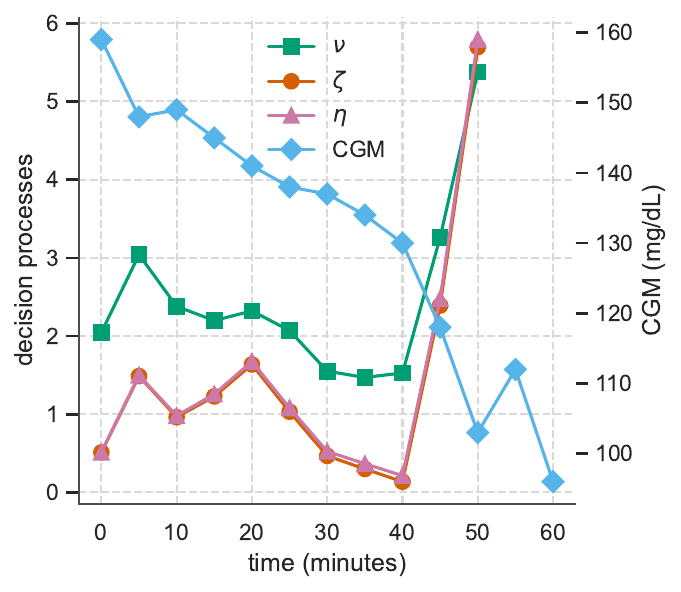}
        \caption{
            The CGM and decision processes during the monitoring window that is shaded in \autoref{fig:cgm_window}.
        }\label{fig:cgm_traj_dive}
    \end{subfigure}\hfill
    \begin{subfigure}{0.475\textwidth}
\includegraphics[width=\linewidth]{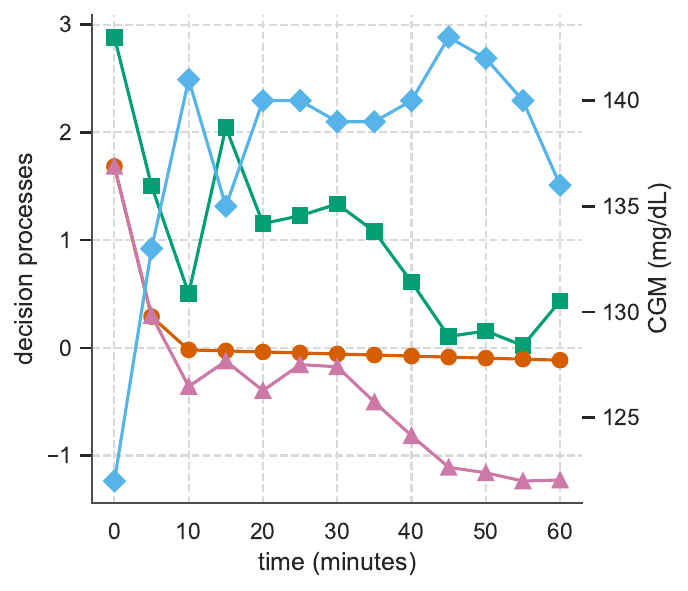}
        \caption{
            Another monitoring window not followed by hypoglycemia.
        }\label{fig:cgm_traj_stable}
    \end{subfigure}
    \caption{
        Example CGM trajectories and corresponding decision processes. The upper panel illustrates the construction of a monitoring window (shaded) and the definition of a hypoglycemia episode. 
        The bottom panels illustrate the decision rule learned under a sensitivity of 0.95 and a cost of 0.7.
        The bottom-left panel shows how the proposed decision rule operates on a window with declining glucose. Once the benefit of an immediate classification ($\zeta$) outweighs that of waiting ($\nu$) at $t = 50$ minutes, the rule issues a hypoglycemia warning based on the positive imminent risk signal ($\eta > 0$). (The $\eta$ and $\zeta$ curves overlap in this panel due to a small Lagrangian multiplier: $\hat a = 0.045$.) The bottom-right panel shows a monitoring window with stable CGM, where the rule continues to wait until the end and issues no warnings.
    }\label{fig:cgm_traj}
\end{figure}

Each sample $\xx[1:T]$ is a one-hour CGM sequence recorded at 5-minute intervals ($T=13$). The binary outcome $\yy$ indicates whether a hypoglycemic event occurs within the subsequent 30 minutes, defined as glucose $\leq60$ mg/dL sustained for at least 20 minutes. The one-hour window itself must contain no hypoglycemia. \autoref{fig:cgm_window} illustrates an example trajectory.

\begin{figure}
    \centering
    \includegraphics[width=\linewidth]{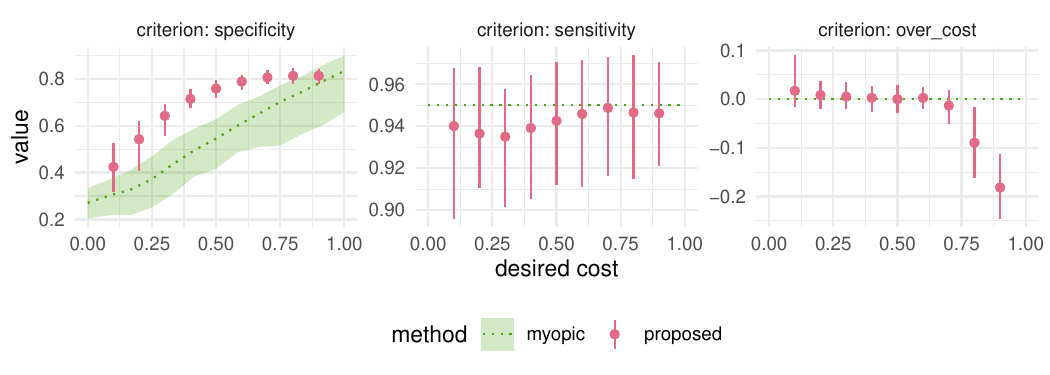}
    \caption{
        Performance of the proposed sequential decision rules and the myopic baseline under a sensitivity constraint of $\beta = 0.95$ across cost $\gamma = 0.1, \dots, 0.9$. 
        The leftmost panel illustrates the specificity--cost trade-off under the sensitivity constraint. The middle and rightmost panels examine whether the constraints were satisfied; the over cost (the rightmost panel) is defined as the achieved cost minus the desired cost.
        Means (dots) and 90\% intervals (bars and ribbons) are shown over 50 repeated cross-validation experiments.
        To reduce visual clutter, the dots and bars for the myopic baseline are replaced with dotted lines and ribbons.
    }\label{fig:cgm_sum}
\end{figure}

\begin{figure}
    \begin{subfigure}{\textwidth}
        \includegraphics[width=\linewidth]{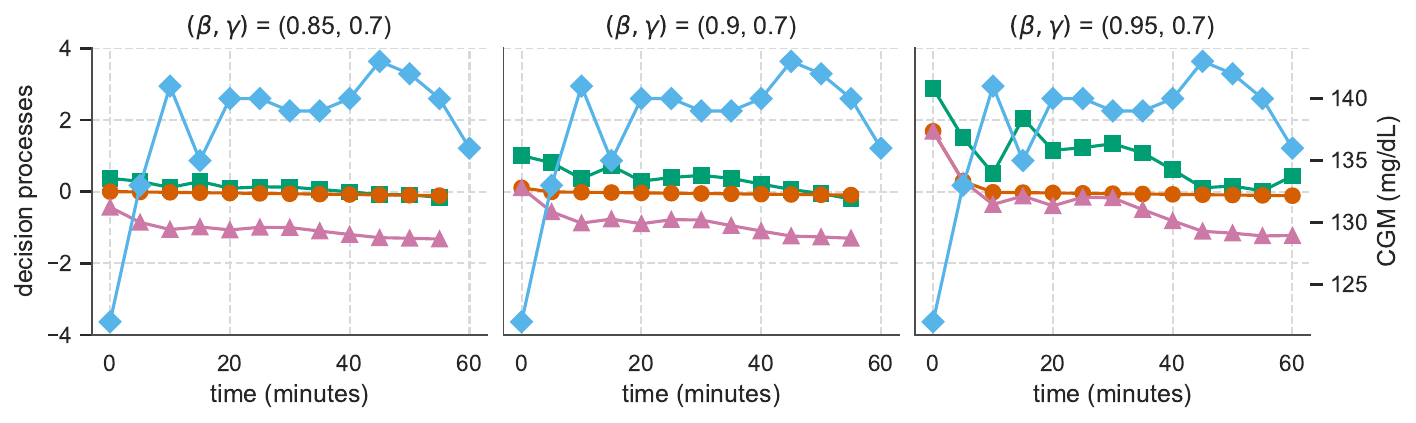}
        \caption{
            Set $\gamma = 0.7$ and vary $\beta = 0.85, 0.9, 0.95$.
        }\label{fig:cgm_traj_stable_multi_tpr}
    \end{subfigure}\\
    \begin{subfigure}{\textwidth}
        \includegraphics[width=\linewidth]{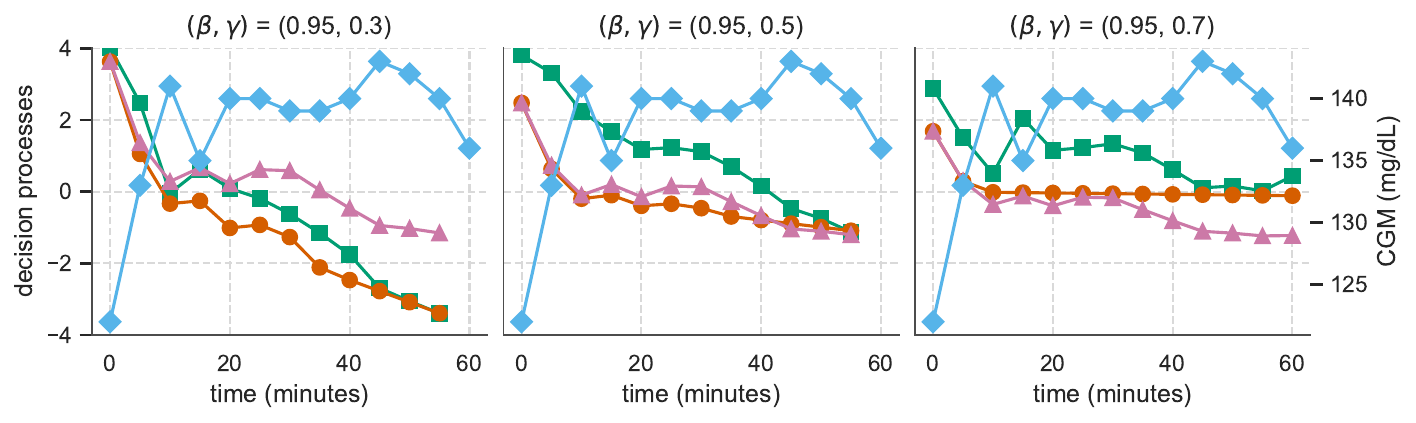}
        \caption{
            Set $\beta = 0.95$ and vary $\gamma = 0.3, 0.5, 0.7$.
        }\label{fig:cgm_traj_stable_multi_cost}
    \end{subfigure}
    \caption{
        Effects of different sensitivity ($\beta$) and cost ($\gamma$) constraints on the decision processes for the non-hypoglycemia CGM trajectory (\autoref{fig:cgm_traj_stable}).
    }\label{fig:cgm_traj_multicons}
\end{figure}

For each experiment, we constructed 8,192 monitoring windows by block-bootstrapping the original CGM traces, up-sampling windows preceding hypoglycemia to achieve an event rate of approximately 0.3.
The data were split into training (70\%), validation (10\%), and testing (20\%) sets, with windows derived from the same hypoglycemia episode kept within the same partition to avoid leakage. The training set was used for learning model parameters, while the validation set was used for performance monitoring and stopping training.
The experiments were repeated 50 times. We estimated sequential rules targeting a high-sensitivity constraint of $\beta = 0.95$, reflecting clinical priorities to avoid missed hypoglycemia.

\autoref{fig:cgm_sum} shows that the CGM setting exhibits the same oracle-like behavior observed in our synthetic experiments. The left panel presents the elbow shape of the specificity--cost trade-off. At the same time, the two rightmost bars show that the sequential rule consistently underuses cost when $\gamma > 0.7$, indicating that the full hour is unnecessary for many monitoring windows. Across experiments, the sequential rule achieved near-maximal specificity ($\approx$ 0.8) at an average cost of around 0.7, corresponding to a lead time of approximately 18 minutes. By contrast, the myopic baseline required the entire one-hour window to reach similar specificity.

To illustrate how the learned rule operates in practice, \autoref{fig:cgm_traj} presents two example CGM trajectories and their corresponding decision processes. In both cases, the decision rule was learned under a sensitivity constraint of 0.95 and a cost level of 0.7.
\autoref{fig:cgm_traj_dive} illustrates a monitoring window with declining glucose that is followed by a hypoglycemia episode. Observing the steep dive in glucose level around 40 to 50 minutes, the benefit of making an immediate classification outweighs that of waiting ($\zeta > \nu$), triggering an early warning.
\autoref{fig:cgm_traj_stable}, in contrast, shows a monitoring window after which no hypoglycemia occurs. Under this stable glucose pattern, the rule continues to wait until the end of the window and issues no warning.
    Moreover, \autoref{fig:cgm_traj_multicons} shows how the decision processes vary under different sensitivity and cost constraints. 
    Overall, $\eta$, the benefit associated with a positive decision, tends to increase with $\beta$, as a stricter sensitivity requirement induces a stronger tendency toward positive conclusions (\autoref{fig:cgm_traj_stable_multi_tpr}).
    Likewise, $\nu$, the benefit associated with continuation, tends to increase with $\gamma$, because a looser constraint on cost or timeliness makes delaying the decision more favorable (\autoref{fig:cgm_traj_stable_multi_cost}).

This example demonstrates that the proposed sequential decision-rule framework can be applied to real-world data beyond synthetic settings. It delivers early, accurate warnings, satisfies stringent sensitivity constraints, and requires substantially fewer observations than myopic rules.

\section{Discussion}\label{sec:diss}

Our contributions are threefold. First, we provide a statistical formulation of optimal sequential classification under a three-fold trade-off among specificity, sensitivity, and monitoring cost. Second, we develop a scalable learning procedure that combines recurrent neural networks with a primal--dual updating scheme to estimate the associated value processes. Finally, we demonstrate through simulations and a CGM application that the method yields interpretable, constraint-satisfying decision rules with strong empirical performance.

Several limitations and directions for further work remain. 
First, although we use flexible recurrent neural network architectures as estimators, we still rely on parametric function classes. Consequently, performance can be sensitive to how well the network models are specified and estimated, as observed in the simulation examples. Developing systematic diagnostic tools and calibration checks to guide model refinement is therefore an important direction for future work. 
Second, we have focused on binary outcomes and a scalar cost structure. Many applications require handling multiple event types, time-to-event outcomes, or richer, patient-specific cost functions that incorporate downstream resource use or harm. Extending the formulation to such settings is a natural next step. We are currently pursuing these extensions.

\begin{acks}[Acknowledgments]
    The authors would like to thank the anonymous referees, the Associate
    Editor and the Editor for their careful review and constructive comments that improved the quality of this paper.
\end{acks}

\begin{supplement}
\stitle{Supplement A: Proofs and Additional Experiment Details}
\sdescription{This supplement contains proofs and additional details for the numeric experiments.}
\end{supplement}

\begin{supplement}
\stitle{Supplement B: Code}
\sdescription{
    Source code implementing the proposed method. 
}
\end{supplement}

\bibliographystyle{imsart-nameyear} \bibliography{references}       

\end{document}